# ScanNet: Richly-annotated 3D Reconstructions of Indoor Scenes

Angela Dai[1]    Angel X. Chang[2]    Manolis Savva[2]    Maciej Halber[2]    Thomas Funkhouser[2]    Matthias Nießner[1,3]
[1]Stanford University    [2]Princeton University    [3]Technical University of Munich
www.scan-net.org

## Abstract

*A key requirement for leveraging supervised deep learning methods is the availability of large, labeled datasets. Unfortunately, in the context of RGB-D scene understanding, very little data is available – current datasets cover a small range of scene views and have limited semantic annotations. To address this issue, we introduce ScanNet, an RGB-D video dataset containing 2.5M views in 1513 scenes annotated with 3D camera poses, surface reconstructions, and semantic segmentations. To collect this data, we designed an easy-to-use and scalable RGB-D capture system that includes automated surface reconstruction and crowdsourced semantic annotation. We show that using this data helps achieve state-of-the-art performance on several 3D scene understanding tasks, including 3D object classification, semantic voxel labeling, and CAD model retrieval.*

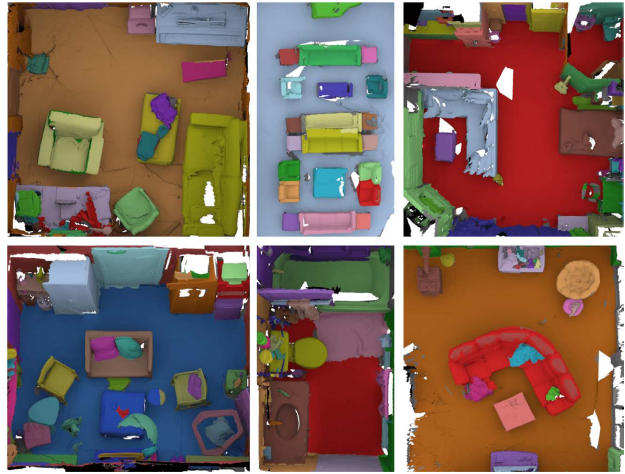

Figure 1. Example reconstructed spaces in ScanNet annotated with instance-level object category labels through our crowdsourced annotation framework.

## 1. Introduction

Since the introduction of commodity RGB-D sensors, such as the Microsoft Kinect, the field of 3D geometry capture has gained significant attention and opened up a wide range of new applications. Although there has been significant effort on 3D reconstruction algorithms, general 3D scene understanding with RGB-D data has only very recently started to become popular. Research along semantic understanding is also heavily facilitated by the rapid progress of modern machine learning methods, such as neural models. One key to successfully applying theses approaches is the availability of large, labeled datasets. While much effort has been made on 2D datasets [17, 44, 47], where images can be downloaded from the web and directly annotated, the situation for 3D data is more challenging. Thus, many of the current RGB-D datasets [74, 92, 77, 32] are orders of magnitude smaller than their 2D counterparts. Typically, 3D deep learning methods use synthetic data to mitigate this lack of real-world data [91, 6].

One of the reasons that current 3D datasets are small is because their capture requires much more effort, and efficiently providing (dense) annotations in 3D is non-trivial. Thus, existing work on 3D datasets often fall back to polygon or bounding box annotations on 2.5D RGB-D images [74, 92, 77], rather than directly annotating in 3D. In the latter case, labels are added manually by expert users (typically by the paper authors) [32, 71] which limits their overall size and scalability.

In this paper, we introduce *ScanNet*, a dataset of richly-annotated RGB-D scans of real-world environments containing 2.5M RGB-D images in 1513 scans acquired in 707 distinct spaces. The sheer magnitude of this dataset is larger than any other [58, 81, 92, 75, 3, 71, 32]. However, what makes it particularly valuable for research in scene understanding is its annotation with estimated calibration parameters, camera poses, 3D surface reconstructions, textured meshes, dense object-level semantic segmentations, and aligned CAD models (see Fig. 2). The semantic segmentations are more than an order of magnitude larger than any previous RGB-D dataset.

In the collection of this dataset, we have considered two main research questions: 1) how can we design a framework that allows many people to collect and annotate large amounts of RGB-D data, and 2) can we use the rich annotations and data quantity provided in ScanNet to learn better 3D models for scene understanding?

| Dataset | Size | Labels | Annotation Tool | Reconstruction | CAD Models |
| --- | --- | --- | --- | --- | --- |
| NYU v2 [58] | 464 scans | 1449 frames | 2D LabelMe-style [69] | none | some [25] |
| TUM [81] | 47 scans | none | - | aligned poses (Vicon) | no |
| SUN 3D [92] | 415 scans | 8 scans | 2D polygons | aligned poses [92] | no |
| SUN RGB-D [75] | 10k frames | 10k frames | 2D polygons + bounding boxes | aligned poses [92] | no |
| BuildingParser [3] | 265 rooms | 265 rooms | CloudCompare [24] | point cloud | no |
| PiGraphs [71] | 26 scans | 26 scans | dense 3D, by the authors [71] | dense 3D [62] | no |
| SceneNN [32] | 100 scans | 100 scans | dense 3D, by the authors [60] | dense 3D [9] | no |
| **ScanNet (ours)** | **1513 scans 2.5M frames** | **1513 scans** | **dense 3D, crowd-sourced MTurk labels also proj. to 2D frames** | **dense 3D [12]** | **yes** |

Table 1. Overview of RGB-D datasets for 3D reconstruction and semantic scene understanding. Note that in addition to the 1513 scans in ScanNet, we also provided dense 3D reconstruction and annotations on all NYU v2 sequences.

To investigate the first question, we built a capture pipeline to help novices acquire semantically-labeled 3D models of scenes. A person uses an app on an iPad mounted with a depth camera to acquire RGB-D video, and then we processes the data off-line and return a complete semantically-labeled 3D reconstruction of the scene. The challenges in developing such a framework are numerous, including how to perform 3D surface reconstruction robustly in a scalable pipeline and how to crowdsource semantic labeling. The paper discusses our study of these issues and documents our experience with scaling up RGB-D scan collection (20 people) and annotation (500 crowd workers).

To investigate the second question, we trained 3D deep networks with the data provided by ScanNet and tested their performance on several scene understanding tasks, including 3D object classification, semantic voxel labeling, and CAD model retrieval. For the semantic voxel labeling task, we introduce a new volumetric CNN architecture.

Overall, the contributions of this paper are:
- A large 3D dataset containing 1513 RGB-D scans of over 707 unique indoor environments with estimated camera parameters, surface reconstructions, textured meshes, semantic segmentations. We also provide CAD model placements for a subset of the scans.
- A design for efficient 3D data capture and annotation suitable for novice users.
- New RGB-D benchmarks and improved results for state-of-the art machine learning methods on 3D object classification, semantic voxel labeling, and CAD model retrieval.
- A complete open source acquisition and annotation framework for dense RGB-D reconstructions.

## 2. Previous Work

A large number of RGB-D datasets have been captured and made publicly available for training and benchmarking [56, 34, 50, 65, 79, 83, 74, 4, 58, 81, 15, 55, 1, 68, 30, 51, 21, 48, 43, 92, 80, 61, 72, 93, 36, 16, 35, 57, 40, 29, 70, 52, 45, 95, 75, 9, 33, 85, 71, 32, 3, 10, 78, 2].[1] These datasets have been used to train models for many 3D scene understanding tasks, including semantic segmentation [67, 58, 26, 86], 3D object detection [73, 46, 27, 76, 77], 3D object classification [91, 53, 66], and others [94, 22, 23].

Most RGB-D datasets contain scans of individual objects. For example, the Redwood dataset [10] contains over 10,000 scans of objects annotated with class labels, 1,781 of which are reconstructed with KinectFusion [59]. Since the objects are scanned in isolation without scene context, the dataset's focus is mainly on evaluating surface reconstruction quality rather than semantic understanding of complete scenes.

One of the earliest and most popular datasets for RGB-D scene understanding is NYU v2 [74]. It is composed of 464 short RGB-D sequences, from which 1449 frames have been annotated with 2D polygons denoting semantic segmentations, as in LabelMe [69]. SUN RGB-D [75] follows up on this work by collecting 10,335 RGB-D frames annotated with polygons in 2D and bounding boxes in 3D. These datasets have scene diversity comparable to ours, but include only a limited range of viewpoints, and do not provide complete 3D surface reconstructions, dense 3D semantic segmentations, or a large set of CAD model alignments.

One of the first RGB-D datasets focused on long RGB-D sequences in indoor environments is SUN3D. It contains a set of 415 Kinect v1 sequences of 254 unique spaces. Although some objects were annotated manually with 2D polygons, and 8 scans have estimated camera poses based on user input, the bulk of the dataset does not include camera poses, 3D reconstructions, or semantic annotations.

Recently, Armeni et al. [3, 2] introduced an indoor dataset containing 3D meshes for 265 rooms captured with a custom Matterport camera and manually labeled with semantic annotations. The dataset is high-quality, but the cap-ture pipeline is based on expensive and less portable hardware. Furthermore, only a fused point cloud is provided as output. Due to the lack of raw color and depth data, its applicability to research on reconstruction and scene understanding from raw RGB-D input is limited.

---

[1] A comprehensive and detailed overview of publicly-accessible RGB-D datasets is given by [20] at http://www0.cs.ucl.ac.uk/staff/M.Firman/RGBDdatasets/, which is updated on a regular basis.

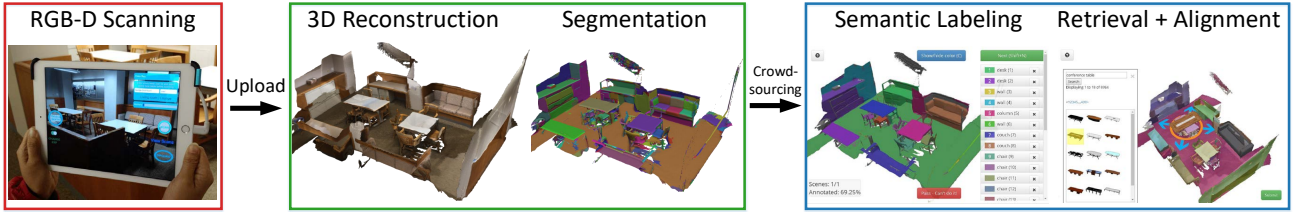

Figure 2. Overview of our RGB-D reconstruction and semantic annotation framework. **Left:** a novice user uses a handheld RGB-D device with our scanning interface to scan an environment. **Mid:** RGB-D sequences are uploaded to a processing server which produces 3D surface mesh reconstructions and their surface segmentations. **Right:** Semantic annotation tasks are issued for crowdsourcing to obtain instance-level object category annotations and 3D CAD model alignments to the reconstruction.

The datasets most similar to ours are SceneNN [32] and PiGraphs [71], which are composed of 100 and 26 densely reconstructed and labeled scenes respectively. The annotations are done directly in 3D [60, 71]. However, both scanning and labeling are performed only by expert users (i.e. the authors), limiting the scalability of the system and the size of the dataset. In contrast, we design our RGB-D acquisition framework specifically for ease-of-use by untrained users and for scalable processing through crowdsourcing. This allows us to acquire a significantly larger dataset with more annotations (currently, 1513 sequences are reconstructed and labeled).

## 3. Dataset Acquisition Framework

In this section, we focus on the design of the framework used to acquire the ScanNet dataset (Fig. 2). We discuss design trade-offs in building the framework and relay findings on which methods were found to work best for large-scale RGB-D data collection and processing.

Our main goal driving the design of our framework was to allow untrained users to capture semantically labeled surfaces of indoor scenes with commodity hardware. Thus the RGB-D scanning system must be trivial to use, the data processing robust and automatic, the semantic annotations crowdsourced, and the flow of data through the system handled by a tracking server.

### 3.1. RGB-D Scanning

**Hardware.** There is a spectrum of choices for RGB-D sensor hardware. Our requirement for deployment to large groups of inexperienced users necessitates a portable and low-cost RGB-D sensor setup. We use the Structure sensor [63], a commodity RGB-D sensor with design similar to the Microsoft Kinect v1. We attach this sensor to a handheld device such as an iPhone or iPad (see Fig. 2 left) — results in this paper were collected using iPad Air2 devices. The iPad RGB camera data is temporally synchronized with the depth sensor via hardware, providing synchronized depth and color capture at 30 Hz. Depth frames are captured at a resolution of $640 \times 480$ and color at $1296 \times 968$ pixels. We enable auto-white balance and auto-exposure by default.

**Calibration.** Our use of commodity RGB-D sensors necessitates unwarping of depth data and alignment of depth and color data. Prior work has focused mostly on controlled lab conditions with more accurate equipment to inform calibration for commodity sensors (e.g., Wang et al. [87]). However, this is not practical for novice users. Thus the user only needs to print out a checkerboard pattern, place it on a large, flat surface, and capture an RGB-D sequence viewing the surface from close to far away. This sequence, as well as a set of infrared and color frame pairs viewing the checkerboard, are uploaded by the user as input to the calibration. Our system then runs a calibration procedure based on [84, 14] to obtain intrinsic parameters for both depth and color sensors, and an extrinsic transformation of depth to color. We find that this calibration procedure is easy for users and results in improved data and consequently enhanced reconstruction quality.

**User Interface.** To make the capture process simple for untrained users, we designed an iOS app with a simple live RGB-D video capture UI (see Fig. 2 left). The user provides a name and scene type for the current scan and proceeds to record a sequence. During scanning, a log-scale RGB feature detector point metric is shown as a "featurefulness" bar to provide a rough measure of tracking robustness and reconstruction quality in different regions being scanned. This feature was critical for providing intuition to users who are not familiar with the constraints and limitations of 3D reconstruction algorithms.

**Storage.** We store scans as compressed RGB-D data on the device flash memory so that a stable internet connection is not required during scanning. The user can upload scans to the processing server when convenient by pressing an "upload" button. Our sensor units used 128 GB iPad Air2 devices, allowing for several hours of recorded RGB-D video. In practice, the bottleneck was battery life rather than storage space. Depth is recorded as 16-bit unsigned short values and stored using standard zLib compression. RGB data is encoded with the H.264 codec with a high bitrate of $15\,\mathrm{Mbps}$ to prevent encoding artifacts. In addition to the RGB-D frames, we also record Inertial Measurement Unit (IMU) data, including acceleration, and angular velocities, from the Apple SDK. Timestamps are recorded for IMU, color, and depth images.

### 3.2. Surface Reconstruction

Once data has been uploaded from the iPad to our server, the first processing step is to estimate a densely-reconstructed 3D surface mesh and 6-DoF camera poses for all RGB-D frames. To conform with the goal for an automated and scalable framework, we choose methods that favor robustness and processing speed such that uploaded recordings can be processed at near real-time rates with little supervision.

**Dense Reconstruction.** We use volumetric fusion [11] to perform the dense reconstruction, since this approach is widely used in the context of commodity RGB-D data. There is a large variety of algorithms targeting this scenario [59, 88, 7, 62, 37, 89, 42, 9, 90, 38, 12]. We chose the BundleFusion system [12] as it was designed and evaluated for similar sensor setups as ours, and provides real-time speed while being reasonably robust given handheld RGB-D video data.

For each input scan, we first run BundleFusion [12] at a voxel resolution of $1\,\mathrm{cm}^3$. BundleFusion produces accurate pose alignments which we then use to perform volumetric integration through VoxelHashing [62] and extract a high resolution surface mesh using the Marching Cubes algorithm on the implicit TSDF ($4\,\mathrm{mm}^3$ voxels). The mesh is then automatically cleaned up with a set of filtering steps to merge close vertices, delete duplicate and isolated mesh parts, and finally to downsample the mesh to high, medium, and low resolution versions (each level reducing the number of faces by a factor of two).

**Orientation.** After the surface mesh is extracted, we automatically align it and all camera poses to a common coordinate frame with the $z$-axis as the up vector, and the $xy$ plane aligned with the floor plane. To perform this alignment, we first extract all planar regions of sufficient size, merge regions defined by the same plane, and sort them by normal (we use a normal threshold of $25°$ and a planar offset threshold of $5\,\mathrm{cm}$). We then determine a prior for the up vector by projecting the IMU gravity vectors of all frames into the coordinates of the first frame. This allows us to select the floor plane based on the scan bounding box and the normal most similar to the IMU up vector direction. Finally, we use a PCA on the mesh vertices to determine the rotation around the $z$-axis and translate the scan such that its bounds are within the positive octant of the coordinate system.

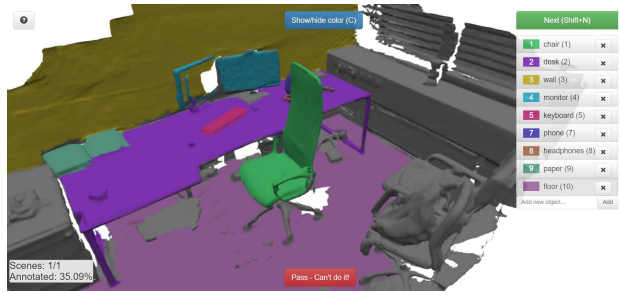

Figure 3. Our web-based crowdsourcing interface for annotating a scene with instance-level object category labels. The right panel lists object instances already annotated in the scene with matching painted colors. This annotation is in progress at $\approx 35\%$, with gray regions indicating unannotated surfaces.

**Validation.** This reconstruction process is automatically triggered when a scan is uploaded to the processing server and runs unsupervised. In order to establish a clean snapshot to construct the ScanNet dataset reported in this paper, we automatically discard scan sequences that are short, have high residual reconstruction error, or have low percentage of aligned frames. We then manually check for and discard reconstructions with noticeable misalignments.

### 3.3. Semantic Annotation

After a reconstruction is produced by the processing server, annotation HITs (Human Intelligence Tasks) are issued on the Amazon Mechanical Turk crowdsourcing market. The two HITs that we crowdsource are: i) instance-level object category labeling of all surfaces in the reconstruction, and ii) 3D CAD model alignment to the reconstruction. These annotations are crowdsourced using web-based interfaces to again maintain the overall scalability of the framework.

**Instance-level Semantic Labeling.** Our first annotation step is to obtain a set of object instance-level labels directly on each reconstructed 3D surface mesh. This is in contrast to much prior work that uses 2D polygon annotations on RGB or RGB-D images, or 3D bounding box annotations.

We developed a WebGL interface that takes as input the low-resolution surface mesh of a given reconstruction and a conservative over-segmentation of the mesh using a normal-based graph cut method [19, 39]. The crowd worker then selects segments to annotate with instance-level object category labels (see Fig. 3). Each worker is required to annotate at least 25% of the surfaces in a reconstruction, and encouraged to annotate more than 50% before submission. Each scan is annotated by multiple workers (scans in ScanNet are annotated by 2.3 workers on average).

A key challenge in designing this interface is to enable efficient annotation by workers who have no prior experience with the task, or 3D interfaces in general. Our interface uses a simple painting metaphor where clicking and drag-ging over surfaces paints segments with a given label and corresponding color. This functions similarly to 2D painting and allows for erasing and modifying existing regions.

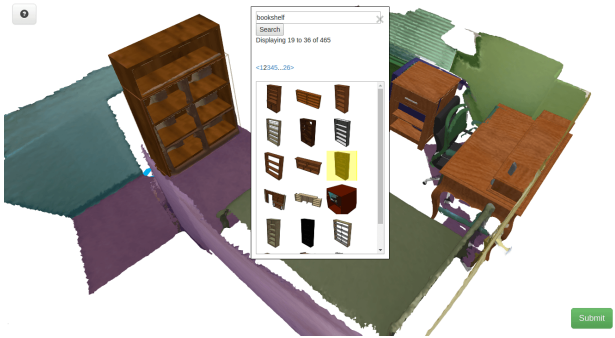

Figure 4. Crowdsourcing interface for aligning CAD models to objects in a reconstruction. Objects can be clicked to initiate an assisted search for CAD models (see list of bookshelves in middle). A suggested model is placed at the position of the clicked object, and the user then refines the position and orientation. A desk, chair, and nightstand have been already placed here.

| Statistic | SceneNN [32] | ScanNet |
|---|---|---|
| # of scans | 100 | 1513 |
| # of RGB-D frames | 2,475,905 | 2,492,518 |
| floor area (avg / sum $m^2$) | 22.6 / 2,124 | 22.6 / 34,453 |
| surface area (avg / sum $m^2$) | 75.3 / 7,078 | 51.6 / 78,595 |
| labeled objects (avg / sum) | 15.8 / 1482 | 24.1 / 36,213 |

Table 2. Summary statistics for ScanNet compared to the most similar existing dataset (SceneNN [32]). ScanNet has an order of magnitude more scans, with 3D surface mesh reconstructions covering more than ten times the floor and surface area, and with more than 36,000 annotated object instances.

Another design requirement is to allow for freeform text labels, to reduce the inherent bias and scalability issues of pre-selected label lists. At the same time, it is desirable to guide users for consistency and coverage of basic object types. To achieve this, the interface provides autocomplete functionality over all labels previously provided by other workers that pass a frequency threshold ($> 5$ annotations). Workers are always allowed to add arbitrary text labels to ensure coverage and allow expansion of the label set.

Several additional design details are important to ensure usability by novice workers. First, a simple distance check for connectedness is used to disallow labeling of disconnected surfaces with the same label. Earlier experiments without this constraint resulted in two undesirable behaviors: cheating by painting many surfaces with a few labels, and labeling of multiple object instances with the same label. Second, the 3D nature of the data is challenging for novice users. Therefore, we first show a full turntable rotation of each reconstruction and instruct workers to change the view using a rotating turntable metaphor. Without the turntable rotation animation, many workers only annotated from the initial view and never used camera controls despite the provided instructions.

**CAD Model Retrieval and Alignment.** In the second annotation task, a crowd worker was given a reconstruction already annotated with object instances and asked to place appropriate 3D CAD models to represent major objects in the scene. The challenge of this task lies in the selection of closely matching 3D models from a large database, and in precisely aligning each model to the 3D position of the corresponding object in the reconstruction.

We implemented an assisted object retrieval interface where clicking on a previously labeled object in a reconstruction immediately searched for CAD models with the same category label in the ShapeNetCore [6] dataset, and placed one example model such that it overlaps with the oriented bounding box of the clicked object (see Fig. 4). The worker then used keyboard and mouse-based controls to adjust the alignment of the model, and was allowed to submit the task once at least three CAD models were placed.

Using this interface, we collected sets of CAD models aligned to each ScanNet reconstruction. Preliminary results indicate that despite the challenging nature of this task, workers select semantically appropriate CAD models to match objects in the reconstructions. The main limitation of this interface is due to the mismatch between the corpus of available CAD models and the objects observed in the ScanNet scans. Despite the diversity of the ShapeNet CAD model dataset (55K objects), it is still hard to find exact instance-level matches for chairs, desks and more rare object categories. A promising way to alleviate this limitation is to algorithmically suggest candidate retrieved and aligned CAD models such that workers can perform an easier verification and adjustment task.

## 4. ScanNet Dataset

In this section, we summarize the data we collected using our framework to establish the ScanNet dataset. This dataset is a snapshot of available data from roughly one month of data acquisition by 20 users at locations in several countries. It has annotations by more than 500 crowd workers on the Mechanical Turk platform. Since the presented framework runs in an unsupervised fashion and people are continuously collecting data, this dataset continues to grow organically. Here, we report some statistics for an initial snapshot of 1513 scans, which are summarized in Table 2.

Fig. 5 plots the distribution of scanned scenes over different types of real-world spaces. ScanNet contains a variety of spaces such as offices, apartments, and bathrooms. The dataset contains a diverse set of spaces ranging from small (e.g., bathrooms, closets, utility rooms) to large (e.g., apartments, classrooms, and libraries). Each scan has been annotated with instance-level semantic category labels through our crowdsourcing task. In total, we deployed 3,391 annotation tasks to annotate all 1513 scans.

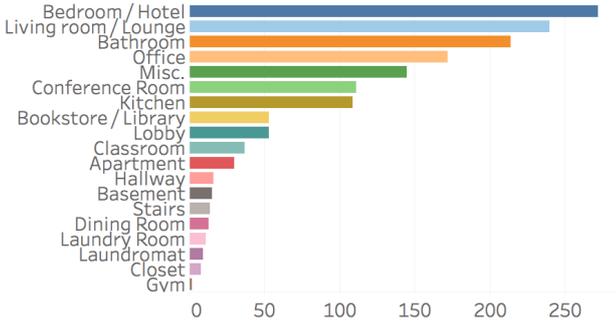

Figure 5. Distribution of the scans in ScanNet organized by type.

|  |  | Scans | | Instances | |
|---|---|---|---|---|---|
|  |  | #Train | #Test | #Train | #Test |
| Object Classification | ScanNet | 1205 | 312 | 9305 | 2606 |
|  | NYU | 452 | 80 | 3260 | 613 |
|  | SceneNN | 70 | 12 | 377 | 66 |
| Semantic Voxel Labeling | ScanNet | 1201 | 312 | 80554 | 21300 |

Table 3. Train/Test split for object classification and dense voxel prediction tasks. Note that the number of instances does not include the rotation augmentation.

The text labels used by crowd workers to annotate object instances are all mapped to the object category sets of NYU v2 [58], ModelNet [91], ShapeNet [6], and WordNet [18] synsets. This mapping is made more robust by a preprocess that collapses the initial text labels through synonym and misspelling detection.

In addition to reconstructing and annotating the 1513 ScanNet scans, we have processed all the NYU v2 RGB-D sequences with our framework. The result is a set of dense reconstructions of the NYU v2 spaces with instance-level object annotations in 3D that are complementary in nature to the existing image-based annotations.

We also deployed the CAD model alignment crowdsourcing task to collect a total of 107 virtual scene interpretations consisting of aligned ShapeNet models placed on a subset of 52 ScanNet scans by 106 workers. There were a total of 681 CAD model instances (of 296 unique models) retrieved and placed on the reconstructions, with an average of 6.4 CAD model instances per annotated scan.

For more detailed statistics on this first ScanNet dataset snapshot, please see the appendix.

## 5. Tasks and Benchmarks

In this section, we describe the three tasks we developed as benchmarks for demonstrating the value of ScanNet data.

**Train/Test split statistics.** Table 3 shows the test and training splits of ScanNet in the context of the object classification and dense voxel prediction benchmarks. Note that our data is significantly larger than any existing comparable dataset. We use these tasks to demonstrate that ScanNet enables the use of deep learning methods for 3D scene understanding tasks with supervised training, and compare performance to that using data from other existing datasets.

### 5.1. 3D Object Classification

With the availability of large-scale synthetic 3D datasets such as [91, 6] and recent advances in 3D deep learning, research has developed approaches to classify objects using only geometric data with volumetric deep nets [91, 82, 52, 13, 66]. All of these methods train on purely synthetic data and focus on isolated objects. Although they show limited evaluation on real-world data, a larger evaluation on realistic scanning data is largely missing. When training data is synthetic and test is performed on real data, there is also a significant discrepancy of test performance, as data characteristics, such as noise and occlusions patterns, are inherently different.

With ScanNet, we close this gap as we have captured a sufficiently large amount of 3D data to use real-world RGB-D input for *both* training and test sets. For this task, we use the bounding boxes of annotated objects in ScanNet, and isolate the contained geometry. As a result, we obtain local volumes around each object instance for which we know the annotated category. The goal of the task is to classify the object represented by a set of scanned points within a given bounding box. For this benchmark, we use 17 categories, with $9,677$ train instances and $2,606$ test instances.

**Network and training.** For object classification, we follow the network architecture of the 3D Network-in-Network of [66], without the multi-orientation pooling step. In order to classify partial data, we add a second channel to the $30^3$ occupancy grid input, indicating known and unknown regions (with 1 and 0, respectively) according to the camera scanning trajectory. As in Qi et al. [66], we use an SGD solver with learning rate 0.01 and momentum 0.9, decaying the learning rate by half every 20 epochs, and training the model for 200 epochs. We augment training samples with 12 instances of different rotations (including both elevation and tilt), resulting in a total training set of $111,660$ samples.

**Benchmark performance.** As a baseline evaluation, we run the 3D CNN approach of Qi et al. [66]. Table 4 shows the performance of 3D shape classification with different train and test sets. The first two columns show results on synthetic test data from ShapeNet [6] including both complete and partial data. Naturally, training with the corresponding synthetic counterparts of ShapeNet provides the best performance, as data characteristics are shared. However, the more interesting case is real-world test data (right-most two columns); here, we show results on test sets of SceneNN [32] and ScanNet. First, we see that training on synthetic data allows only for limited knowledge transfer (first two rows). Second, although the relatively small SceneNN dataset is able to learn within its own dataset to a reasonable degree, it does not generalize to the larger variety of environments found in ScanNet. On the other hand, training on ScanNet translates well to testing on SceneNN; as a result, the test results on SceneNN are significantly improved by using the training data from ScanNet. Interestingly enough, these results can be slightly improved when mixing training data of ScanNet with partial scans of ShapeNet (last row).

|  | Synthetic Test Sets | | Real Test Sets | |
|---|---|---|---|---|
| Training Set | ShapeNet | ShapeNet Partial | SceneNN | ScanNet |
| ShapeNet | **92.5** | 37.6 | 68.2 | 39.5 |
| ShapeNet Partial | 88.5 | **92.1** | 72.7 | 45.7 |
| SceneNN | 19.9 | 27.7 | 69.8 | 48.2 |
| NYU | 26.2 | 26.6 | 72.7 | 53.2 |
| ScanNet | 21.4 | 31.0 | 78.8 | 74.9 |
| ScanNet +ShapeNet Par. | 79.7 | 89.8 | **81.2** | **76.6** |

Table 4. 3D object classification benchmark performance. Percentages give the classification accuracy over all models in each test set (average instance accuracy).

|  | Retrieval from ShapeNet | |
|---|---|---|
| Train | Top 1 NN | Top 3 NNs |
| ShapeNet | 10.4% | 8.0% |
| ScanNet | 12.7% | 11.7% |
| ShapeNet + ScanNet | **77.5%** | **77.0%** |

Table 6. 3D model retrieval benchmark performance. Nearest neighbor models are retrieved for ScanNet objects from ShapeNet-Core. Percentages indicate average instance accuracy of retrieved model to query region.

## 5.2. Semantic Voxel Labeling

A common task on RGB data is semantic segmentation (i.e. labeling pixels with semantic classes) [49]. With our data, we can extend this task to 3D, where the goal is to predict the semantic object label on a per-voxel basis. This task of predicting a semantic class for each visible 3D voxel has been addressed by some prior work, but using handcrafted features to predict a small number of classes [41, 86], or focusing on outdoor environments [8, 5].

**Data Generation.** We first voxelize a scene and obtain a dense voxel grid with $2cm^3$ voxels, where every voxel stores its TSDF value and object class annotation (empty space and unlabeled surface points have their own respective classes). We now extract subvolumes of the scene volume, of dimension $2 \times 31 \times 31 \times 62$ and spatial extent 1.5m × 1.5m × 3m; i.e., a voxel size of $\approx 4.8cm^3$; the two channels represent the occupancy and known/unknown space according to the camera trajectory. These sample volumes are aligned with the $xy$-ground plane. For ground truth data generation, voxel labels are propagated from the scene voxelization to these sample volumes. The samples are chosen that $\geq 2\%$ of the voxels are occupied (i.e., on the surface), and $\geq 70\%$ of these surface voxels have valid annotations; samples not meeting these criteria are discarded. Across ScanNet, we generate $93,721$ subvolume examples for training, augmented by 8 rotations each (i.e., $749,768$ training samples), from $1201$ training scenes. In addition, we extract $18,750$ sample volumes for testing, which are also augmented by 8 rotations each (i.e., $150,000$ test samples) from $312$ test scenes. We have 20 object class labels plus 1 class for free space.

**Network and training.** For the semantic voxel labeling task, we propose a network which predicts class labels for a column of voxels in a scene according to the occupancy characteristics of the voxels' neighborhood. In order to infer labels for an entire scene, we use the network to predict a label for every voxel column at test time (i.e., every $xy$ position that has voxels on the surface). The network takes as input a $2 \times 31 \times 31 \times 62$ volume and uses a series of fully convolutional layers to simultaneously predict class scores for the center column of 62 voxels. We use ReLU and batch normalization for all layers (except the last) in the network. To account for the unbalanced training data over the class labels, we weight the cross entropy loss with the inverse log of the histogram of the train data.

We use an SGD solver with learning rate $0.01$ and momentum $0.9$, decaying the learning rate by half every 20 epochs, and train the model for 100 epochs.

**Quantitative Results.** The goal of this task is to predict semantic labels for all visible surface voxels in a given 3D scene; i.e., every voxel on a visible surface receives one of the 20 object class labels. We use NYU2 labels, and list voxel classification results on ScanNet in Table 7. We achieve an voxel classification accuracy of 73.0% over the set of 312 test scenes, which is based purely on the geometric input (no color is used).

In Table 5, we show our semantic voxel labeling results on the NYU2 dataset [58]. We are able to outperform previous methods which are trained on limited sets of real-world data using our volumetric classification network. For instance, Hermans et al. [31] classify RGB-D frames using a dense random decision forest in combination with a conditional random field. Additionally, SemanticFusion [54] uses a deep net trained on RGB-D frames, and regularize the predictions with a CRF over a 3D reconstruction of the frames; note that we compare to their classification results before the CRF regularization. SceneNet trains on a large synthetic dataset and fine-tunes on NYU2. Note that in contrast to Hermans et al. and SemanticFusion, neither we nor SceneNet use RGB information.

| | floor | wall | chair | table | window | bed | sofa | tv | objs. | furn. | ceil. | avg. |
|---|---|---|---|---|---|---|---|---|---|---|---|---|
| Hermans et al. [31] | 91.5 | 71.8 | 41.9 | 27.7 | 46.1 | 68.4 | 28.5 | **38.4** | 8.6 | 37.1 | **83.4** | 49.4 |
| SemanticFusion [54]* | 92.6 | **86.0** | 58.4 | 34.0 | 60.5 | 61.7 | 47.3 | 33.9 | **59.1** | 63.7 | 43.4 | 58.2 |
| SceneNet [28] | 96.2 | 85.3 | 61.0 | 43.8 | 30.0 | 72.5 | 62.8 | 19.4 | 50.0 | 60.4 | 74.1 | 59.6 |
| Ours (ScanNet + NYU) | **99.0** | 55.8 | **67.6** | **50.9** | **63.1** | **81.4** | **67.2** | 35.8 | 34.6 | **65.6** | 46.2 | **60.7** |

Table 5. Dense pixel classification accuracy on NYU2 [58]. Note that both SemanticFusion [54] and Hermans et. al. [31] use both geometry and color, and that Hermans et al. uses a CRF, unlike our approach which is *geometry-only* and has only unary predictions. The reported SemanticFusion classification is on the 13 class task (13 class average accuracy of 58.9%).

| Class | % of Test Scenes | Accuracy |
|---|---|---|
| Floor | 35.7% | 90.3% |
| Wall | 38.8% | 70.1% |
| Chair | 3.8% | 69.3% |
| Sofa | 2.5% | 75.7% |
| Table | 3.3% | 68.4% |
| Door | 2.2% | 48.9% |
| Cabinet | 2.4% | 49.8% |
| Bed | 2.0% | 62.4% |
| Desk | 1.7% | 36.8% |
| Toilet | 0.2% | 69.9% |
| Sink | 0.2% | 39.4% |
| Window | 0.4% | 20.1% |
| Picture | 0.2% | 3.4% |
| Bookshelf | 1.6% | 64.6% |
| Curtain | 0.7% | 7.0% |
| Shower Curtain | 0.04% | 46.8% |
| Counter | 0.6% | 32.1% |
| Refrigerator | 0.3% | 66.4% |
| Bathtub | 0.2% | 74.3% |
| OtherFurniture | 2.9% | 19.5% |
| Total | - | 73.0% |

Table 7. Semantic voxel label prediction accuracy on ScanNet test scenes.

Note that we do not explicitly enforce prediction consistency between neighboring voxel columns when the *test volume* is slid across the $xy$ plane. This could be achieved with a volumetric CRF [64], as used in [86]; however, our goal in this task to focus exclusively on the per-voxel classification accuracy.

### 5.3. 3D Object Retrieval

Another important task is retrieval of similar CAD models given (potentially partial) RGB-D scans. To this end, one wants to learn a shape embedding where a feature descriptor defines geometric similarity between shapes. The core idea is to train a network on a shape classification task where a shape embedding can be learned as *byproduct* of the classification task. For instance, Wu et al. [91] and Qi et al. [66] use this technique to perform shape retrieval queries within the ShapeNet database.

With ScanNet, we have established category-level correspondences between real-world objects and ShapeNet models. This allows us to train on a classification problem where both real and synthetic data are mixed inside of each category using real and synthetic data within shared class labels. Thus, we can learn an embedding between real and synthetic data in order to perform model retrieval for RGB-D scans. To this end, we use the volumetric shape classification network by Qi et al. [66], we use the same training procedure as in Sec. 5.1. Nearest neighbors are retrieved based on the $\ell_2$ distance between the extracted feature descriptors, and measured against the ground truth provided by the CAD model retrieval task. In Table 6, we show object retrieval results using objects from ScanNet to query for nearest neighbor models from ShapeNetCore. Note that training on ShapeNet and ScanNet independently results in poor retrieval performance, as neither are able to bridge the gap between the differing characteristics of synthetic and real-world data. Training on both ShapeNet and ScanNet together is able to find an embedding of shape similarities between both data modalities, resulting in much higher retrieval accuracy.

## 6. Conclusion

This paper introduces ScanNet: a large-scale RGB-D dataset of 1513 scans with surface reconstructions, instance-level object category annotations, and 3D CAD model placements. To make the collection of this data possible, we designed a scalable RGB-D acquisition and semantic annotation framework that we provide for the benefit of the community. We demonstrated that the richly-annotated scan data collected so far in ScanNet is useful in achieving state-of-the-art performance on several 3D scene understanding tasks; we hope that ScanNet will inspire future work on many other tasks.

## Acknowledgments

This project is funded by Google Tango, Intel, NSF (IIS-1251217 and VEC 1539014/1539099), and a Stanford Graduate fellowship. We also thank Occipital for donating structure sensors and Nvidia for hardware donations, as well as support by the Max-Planck Center for Visual Computing and the Stanford CURIS program. Further, we thank Toan Vuong, Joseph Chang, and Helen Jiang for help on the mobile scanning app and the scanning process, and Hope Casey-Allen and Duc Nugyen for early prototypes of the annotation interfaces. Last but not least, we would like to thank all the volunteers who helped with scanning and get-ting us access to scanning spaces.

## A. Dataset Statistics and Comparisons

In this section, we provide thorough statistics on the construction and composition of ScanNet dataset, and also compare it to the most similar datasets from prior work.

### A.1. Example Annotated Reconstructions

Fig. 6 shows six example annotated reconstructions for a variety of spaces. For each reconstruction, the surface mesh with colors is shown, as well as a visualization with category labels for each object collected using our crowd-sourced annotation interface. Category labels are consistent between spaces and are mapped to WordNet [18] synsets. In addition to the category label, separate object instance labels are also available to indicate multiple instances of a given category, such as distinct chairs around a conference table in the fourth row of Fig. 6.

Fig. 7 shows a larger set of reconstructed spaces in ScanNet to illustrate the variety of spaces that are part of the dataset. The scans range from small spaces with just a few objects (e.g., toilets), to large areas with dozens of objects (e.g., classrooms and studio apartments).

### A.2. Dataset Construction Statistics

The construction of ScanNet was carried out with the RGB-D acquisition and annotation framework described in the main paper. In order to provide an intuition of the scalability of our framework, we report timing statistics for both the reconstruction and annotation steps. The median reconstruction processing time (including data conversion, dense voxel fusion, surface mesh extraction, align-ment, cleanup, and preview thumbnail image rendering) is 11.3 min for each scene. A few outliers exist with significantly higher processing times (on the order of hours), due to unplanned processing server downtime during our data collection (mainly software updates), resulting in a higher mean reconstruction time of 14.9 min.

| ScanNet | | SceneNN [32] | |
|---|---|---|---|
| Category | Count | Category | Count |
| wall | 6226 | chair | 194 |
| chair | 4279 | table | 53 |
| floor | 3212 | floor | 44 |
| table | 2223 | seat | 41 |
| door | 1181 | desk | 39 |
| couch | 1048 | monitor | 31 |
| cabinet | 937 | sofa | 25 |
| desk | 733 | cabinet | 25 |
| shelf | 732 | door | 24 |
| bed | 699 | box | 23 |
| office chair | 669 | keyboard | 23 |
| trashcan | 561 | trash bin | 21 |
| pillow | 490 | wall | 20 |
| sink | 470 | pillow | 19 |
| window | 398 | fridge | 18 |
| toilet | 397 | stand | 18 |
| picture | 351 | bag | 17 |
| bookshelf | 328 | bed | 16 |
| monitor | 308 | window | 14 |
| curtain | 280 | sink | 13 |
| computer | 274 | printer | 12 |
| armchair | 264 | computer | 12 |
| bathtub | 253 | chair01 | 12 |
| coffee table | 239 | desk1 | 11 |
| box | 231 | monitor01 | 10 |
| dining chair | 230 | shelves | 10 |
| refrigerator | 226 | shelf | 10 |
| book | 221 | chair1 | 10 |
| lamp | 218 | chair02 | 10 |
| towel | 216 | fan | 9 |
| kitchen cabinet | 203 | basket | 9 |
| drawer | 202 | desk2 | 9 |
| tv | 187 | laptop | 9 |
| nightstand | 182 | trashbin | 9 |
| counter | 179 | kettle | 9 |
| dresser | 177 | microwave | 9 |
| clothes | 164 | monitor1 | 8 |
| countertop | 163 | stove | 8 |
| stool | 130 | chair2 | 8 |
| plant | 130 | bike | 7 |
| cushion | 116 | blanket | 7 |
| ceiling | 114 | drawer | 7 |
| bedframe | 111 | lamp | 7 |
| keyboard | 107 | wall02 | 7 |
| end table | 105 | wall01 | 7 |
| toilet paper | 104 | wall04 | 7 |
| bag | 104 | backpack | 7 |
| backpack | 100 | cup | 7 |
| blanket | 94 | chair3 | 7 |
| dining table | 94 | whiteboard | 7 |

Table 8. Total counts of annotated object instances of the 50 largest categories in ScanNet (left), and in SceneNN [32] (right), the most similar annotated RGB-D reconstruction dataset. ScanNet contains far more annotated object instances, and the annotated labels are processed for consistency to remove duplicates such as "chair01" in SceneNN.

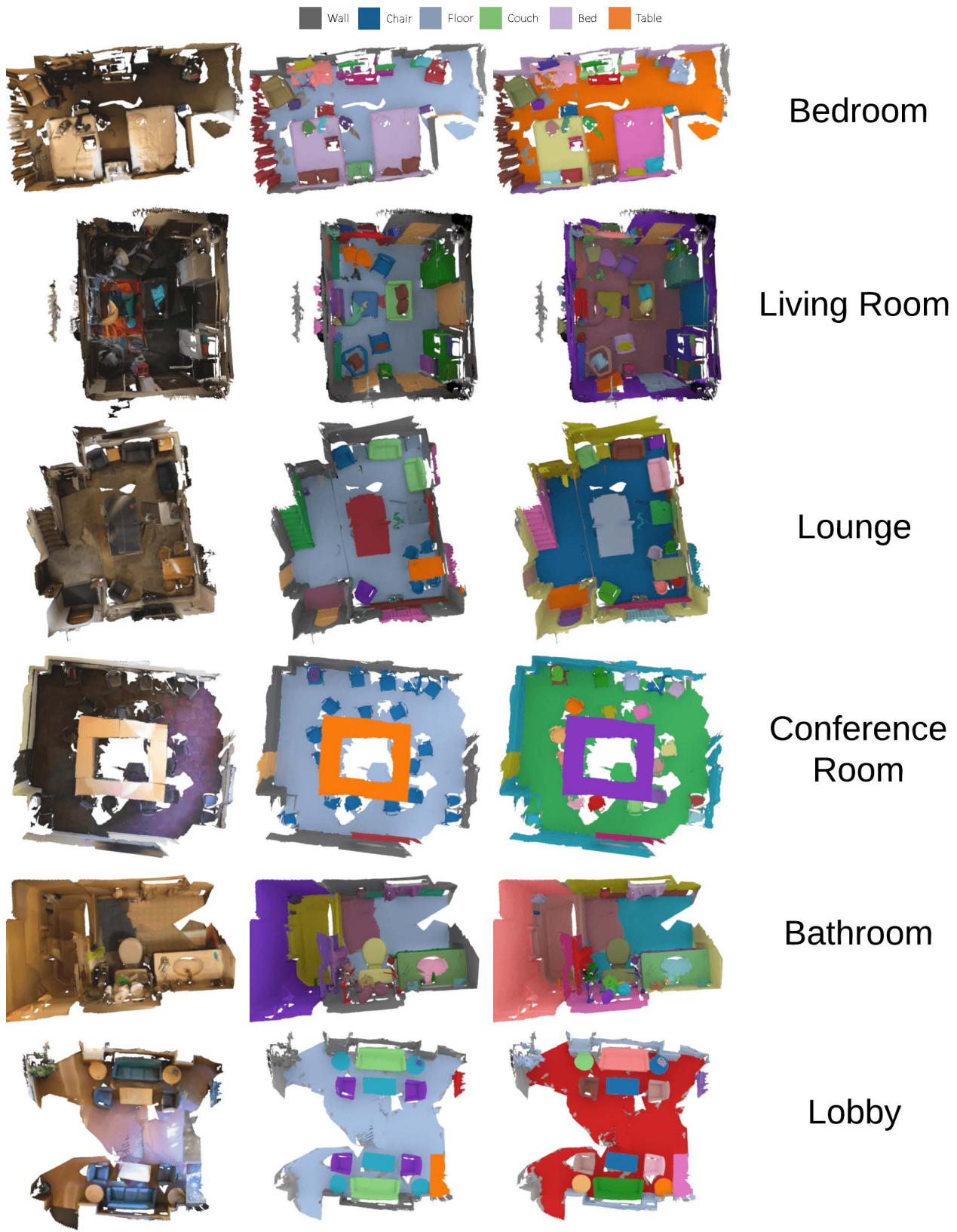

Figure 6. Example annotated scans in ScanNet. **Left:** reconstructed surface mesh with original colors. **Middle:** color indicates category label consistently across all scans. **Right:** each object instance shown with a different randomly assigned color.

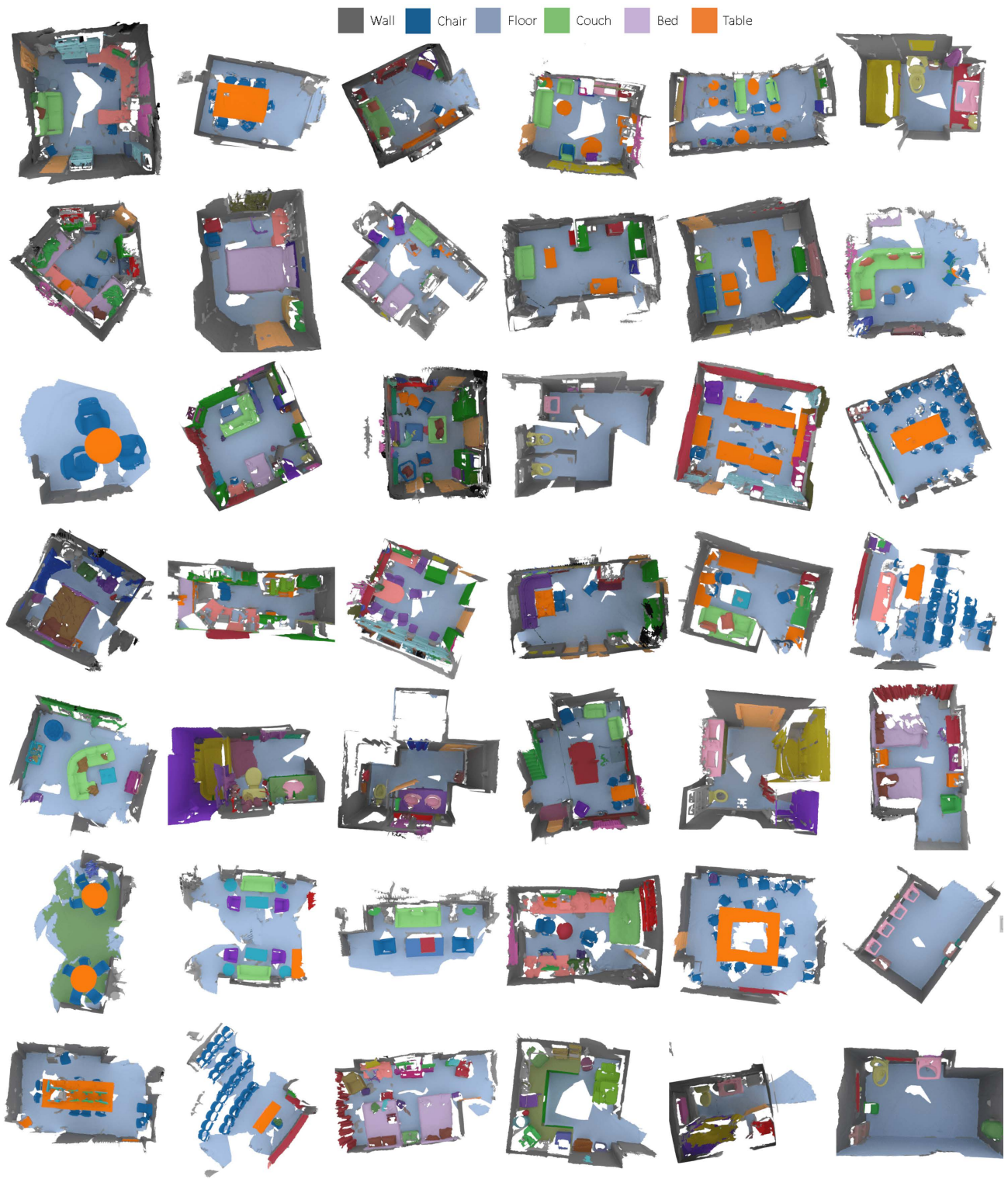

Figure 7. A variety of example annotated scans in ScanNet. Colors indicate category consistently across all scans.

After reconstruction is complete, each scan is annotated by several crowd workers on Amazon Mechanical Turk (2.3 workers on average per scan). The median annotation time per crowd worker is 12.0 min (mean time is 17.3 min, again due to a few outlier workers who take significantly longer). Aggregating the time taken across workers for annotating each of the 1513 scans in ScanNet, the median time per scan is 16.8 min, and the mean time per scan is 22.3 min.

### A.3. Dataset Composition Statistics

The construction of the ScanNet dataset is motivated by the lack of large, annotated, densely reconstructed RGB-D dataset of 3D scenes that are publicly available in the academic community. Existing RGB-D datasets either have full scene-level annotations only for a subset of RGB-D frames (e.g., NYU v2 depth [58]), or they focus on annotating decontextualized objects and not scenes (e.g., Choi et al. [10]). The two datasets that do annotate densely reconstructed RGB-D spaces at the scene level are the SceneNN dataset by Hua et al. [32] and the smaller PiGraphs dataset by Savva et al. [71].

SceneNN consists of 94 RGB-D scans captured using Asus Xtion Pro devices and reconstructed with the method of Choi et al. [9]. The resulting densely-fused surface meshes are fully segmented at the level of meaningful objects. However, only a small set of segments are annotated with semantic labels. On the other hand, the PiGraphs [71] dataset consists of 26 RGB-D scans captured with Kinect v1 devices and reconstructed with the VoxelHashing approach of Nießner et al. [62]. This dataset has more complete and clean semantic labels, including object parts and object instances. However, it contains very few scenes and is limited in the variety of environments, consisting mostly of offices and conference rooms. To illustrate the large gap in quantity of annotated semantic labels between these two datasets and ScanNet, Fig. 8 plots histograms of the total number of labeled object instances and the total numbers of unique semantic labels for each scan.

In order to demonstrate how our category labels map to other data, we plot the distribution of annotated object labels corresponded to the ShapeNetCore 3D CAD model categories in Fig. 9. This mapping is leveraged during our CAD model alignment and retrieval task to automatically suggest instances of CAD models from ShapeNet that match the label of a given object category in the reconstruction.

We can also obtain 2D annotations on the input RGB-D

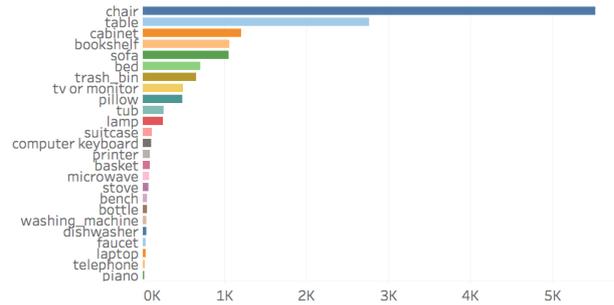

Figure 9. Top 25 most frequent annotation labels in ScanNet scans mapped to ShapeNetCore classes. ScanNet has thousands of 3D reconstructed instances of common objects such as chairs, tables, and cabinets.

sequences by projecting our 3D annotations into each frame using the corresponding camera pose. This way, we obtain an average of 76% annotation coverage of all pixels per scene by using the previously obtained 3D annotations.

### A.4. NYUv2 Reconstruction and Comparison

Here, we discuss how ScanNet relates to NYUv2, one of the most popular RGB-D dataset with annotations. In order to compare the data in ScanNet with the data in NYUv2, we reconstructed and annotated all the RGB-D sequences in NYUv2 using our framework. (Note that for 9 sequences of the NYUv2 dataset, our framework did not obtain valid camera poses for $>50\%$ of the frames, so we did not compute reconstructions and annotations for these sequences.) Moreover, we created a set of surface mesh semantic annotations for the NYUv2 reconstructions by projecting every pixel of the annotated RGB-D frames with valid depth and label into world space using our computed camera poses, and assigning the corresponding object label to the closest surface mesh vertices (within $0.04cm$, using a kd-tree lookup).

We then compare the total surface area of the reconstructed meshes that was annotated using projection from the annotated NYUv2 frames, and using our annotation pipeline. Fig. 10 plots the percentage of reconstructed surfaces in NYUv2 that were annotated with each approach, as well as the percentage distribution for the ScanNet reconstructions for comparison. Note that we exclude the 9 sequences for which we do not have enough valid camera poses.

A noticeable difference between the RGB-D sequences in NYUv2 and those in ScanNet is that overall, the ScanNet sequences are more complete surface reconstructions of the real-world spaces. Most importantly, the NYUv2 original frames in general do not cover a sufficient number of viewpoints of the space to ensure full reconstruction of semantically meaningful complete objects. Fig. 11 shows a comparison of several reconstructed scenes from NYUv2 RGB-D sequences vs comparable reconstructions from ScanNet. As shown in the top-down views, the NYU reconstructions are much more sparse than the ScanNet reconstructions. This disparity makes a more direct comparison with ScanNet reconstructions hard to quantify. However, we can conclude that projecting the annotated NYUv2 RGB-D frames to reconstructions is not sufficient to semantically annotate the spaces, as is clear from the far lower surface coverage distribution for NYUv2 in Fig. 10.

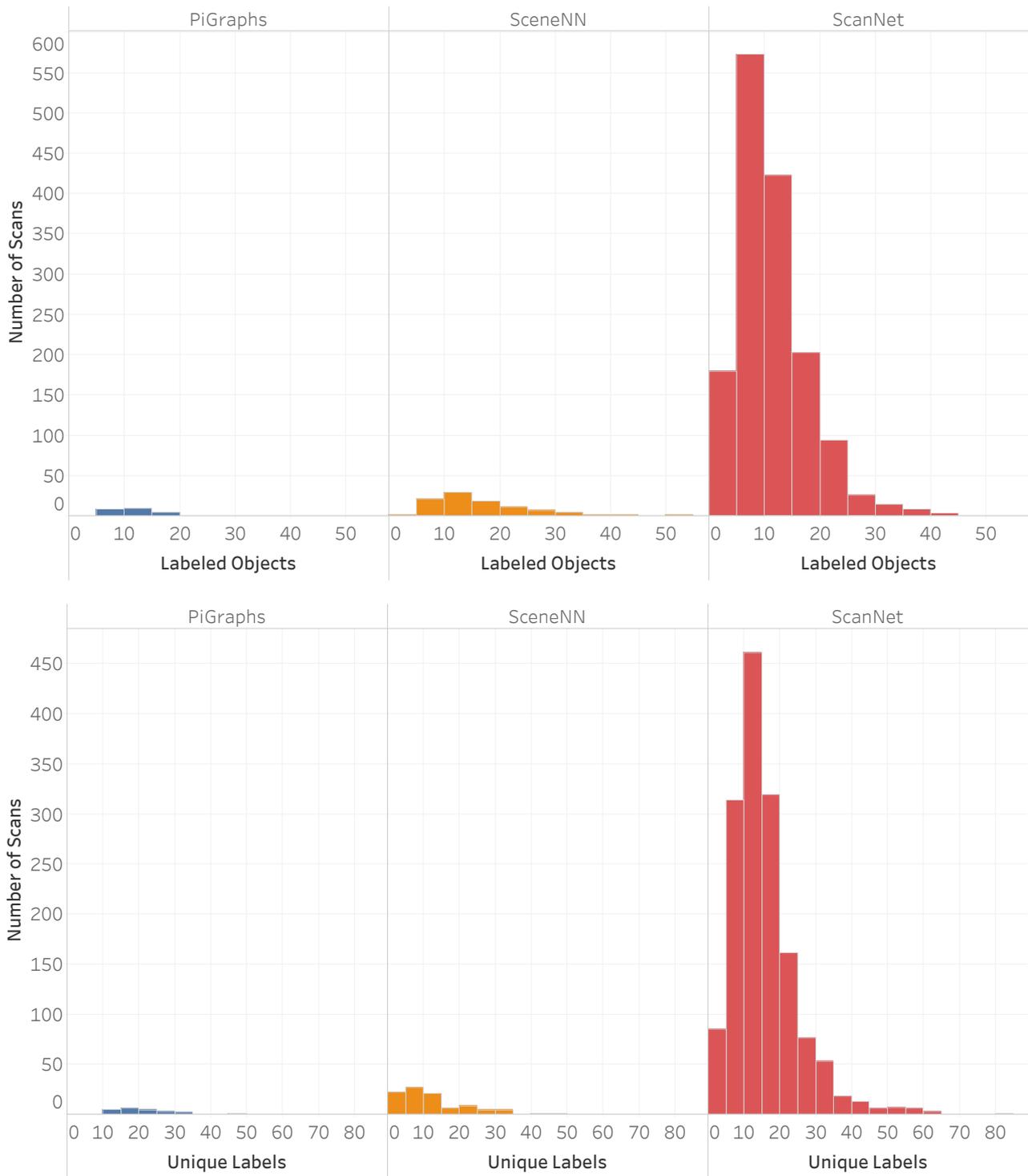

Figure 8. Histograms of the total number of objects labeled per scan (**top**) and total number of unique labels per scan (**bottom**) in the PiGraphs [71], SceneNN [32] and our dataset (ScanNet). The histograms show that ScanNet has many annotated objects over a larger number of scans, ranging in complexity with regards to the total number of objets per scan.

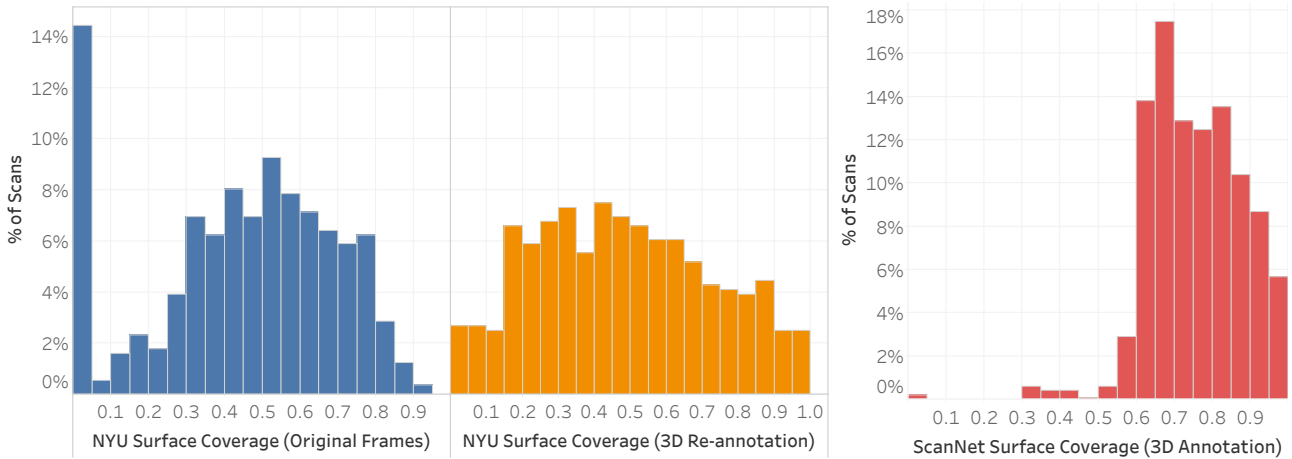

Figure 10. Histograms of the percentage of total reconstruction surface area per scan that is semantically labeled for: NYU v2 reconstructions using projection of RGB-D annotated frames (**left**), for NYU v2 reconstructions using our 3D annotation interface (**middle**), and for ScanNet reconstructions similarly annotated with our interface (**right**).

## B. Tasks

Here we provide more details about the 3D scene understanding tasks and benchmarks discussed in the main paper.

### B.1. Semantic Voxel Labeling

For the semantic voxel labeling task, we propose a network which predicts class labels for each column of a voxelized scene. As shown in Fig. 12, our network takes as input a $2 \times 31 \times 31 \times 62$ volume and uses a series of fully convolutional layers to simultaneously predict class scores for the center column of 62 voxels. We leverage information from the voxel neighborhood of both occupied space (voxels on the surface) and known space (voxels in front of a surface according to the camera trajectory) to describe the input partial data from a scan.

At test time, we slide the network through a scan through a voxelized scan along the $xy$-plane, and each column is predicted independently. Fig. 13 visualizes several ScanNet test scans with voxel label predictions, alongside the ground truth annotations from our crowdsourced labeling task.

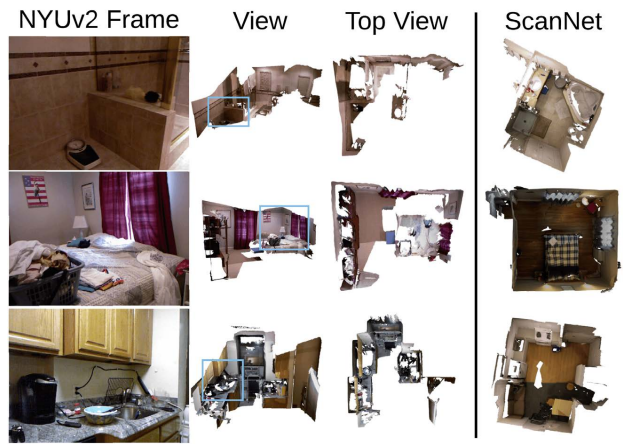

Figure 11. Comparison of reconstructed Bathroom (**top**), Bedroom (**middle**), and Kitchen (**bottom**) from NYUv2 RGB-D frames (**left**), and a comparable reconstruction from ScanNet (**right**). For each NYU scene, we show an example color frame, the rough corresponding region of the view in the reconstructed scene (light blue box), and a top down view of the reconstruction. While NYUv2 reconstruction look complete from some viewpoints, much of the scene is left uncovered (see top down views). In contrast, ScanNet reconstruction have a much more complete coverage of the space and allow for denser annotation.

## C. Dataset Acquisition Framework

This section provides more details for specific steps in our RGB-D data acquisition framework which was described in the main paper. To enable scalable dataset acquisition, we designed our data acquisition framework for 1) ease of use during capture, 2) robust reconstruction, 3) rapid crowdsourcing, 4) visibility into the collected data and its metadata. For 1) we developed an iPad app (see Ap-pendix C.1) with an easy-to-use interface, reasonable scanning presets, and minimalistic user controls. To ensure good reconstruction with minimal user interaction during scanning, we tested different exposure time settings and enabled auto white balancing (see Appendix C.1). We also established a simple calibration process that novice users could carry out (see Appendix C.2), and offloaded RGB-D reconstruction to the cloud (see Appendix C.3). Finally, we developed web-based UIs for crowdsourcing semantic annotation tasks as described in Appendix C.4, and for managing the collected data as described in Appendix C.6.

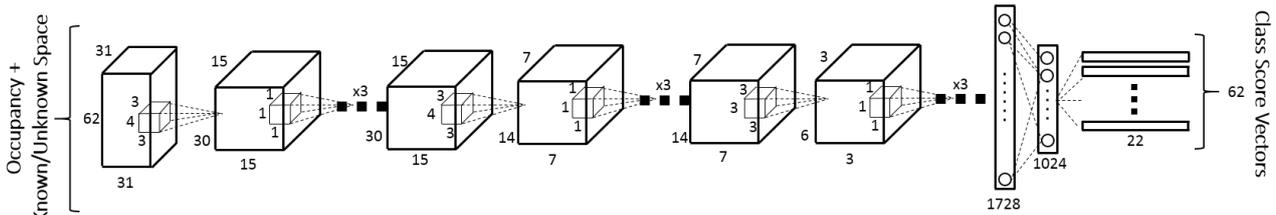

Figure 12. Deep Neural Network architecture for our semantic voxel label prediction task. The network is mainly composed of 3D convolutions that process the geometry of a scene using a 3D voxel grid representation.

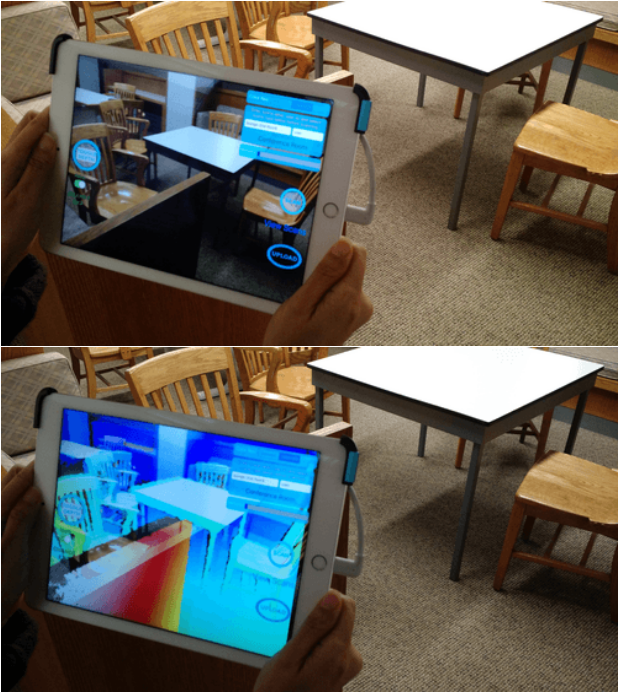

Figure 14. Our RGB-D recording app on an iPad Air2 with attached Structure sensor (showing color stream at the top and depth stream at the bottom). The app allows novice users to record RGB-D videos and upload to a server for reconstruction and annotation.

## C.1. RGB-D Acquisition UI

Fig. 14 shows our RGB-D recording app on the iPad. We designed an iPad app with a simple camera-based UI and a minimalistic set of controls. Before scanning, the user enters a user name, a scene name, and selects the type of room being scanned. The user then presses a single button to start and stop a scan recording. The interface can be toggled between visualizing the color stream and the depth stream overlaid on the color.

We found that the most challenging part of scanning for novice users was acquiring an intuition as to what regions during scanning are likely to result in poor tracking and failed reconstruction. To alleviate this, we added a "progress bar"-style visualization during active scanning which indicates the featurefulness of the region being scanned. The bar ranges from full green, indicating high feature count, to near-empty black, indicating low feature count and high likelihood of tracking loss. This UI element was helpful for quickly familiarizing users with the scanning process. After scanning, the user can view a list of scans on the device and select to upload the scan data to a processing server. During upload, a progress bar is shown and scanning is disabled. Upon completion of the upload, the checksums of scan data on the server are verified against local data and the scans are automatically deleted to provide more memory for scanning.

**Auto white balancing and Exposure Settings** Another challenge towards performing reconstruction in uncontrolled scenarios is the wide variety of illumination conditions. Since our scanning app was designed for novice users, we opted to provide a reasonable set of presets and allow for manual override only when deemed necessary. By default, we enabled continuous automatic whitepoint balancing as implemented by the iOS SDK. We also enabled dynamic exposure selection again as implemented by the iOS SDK, but instructed users that they could manually adjust exposure if necessary to make overly dark locations brighter, or overly bright locations darker. The exposure setting can have a significant impact on the amount of motion blur during scanning. However, we found that inexperienced users preferred to rely on dynamic exposure, and typically moved relatively slowly during scanning, making motion blur less of an issue. The average exposure time during scans with dynamic exposure was close to $30\,\text{ms}$.

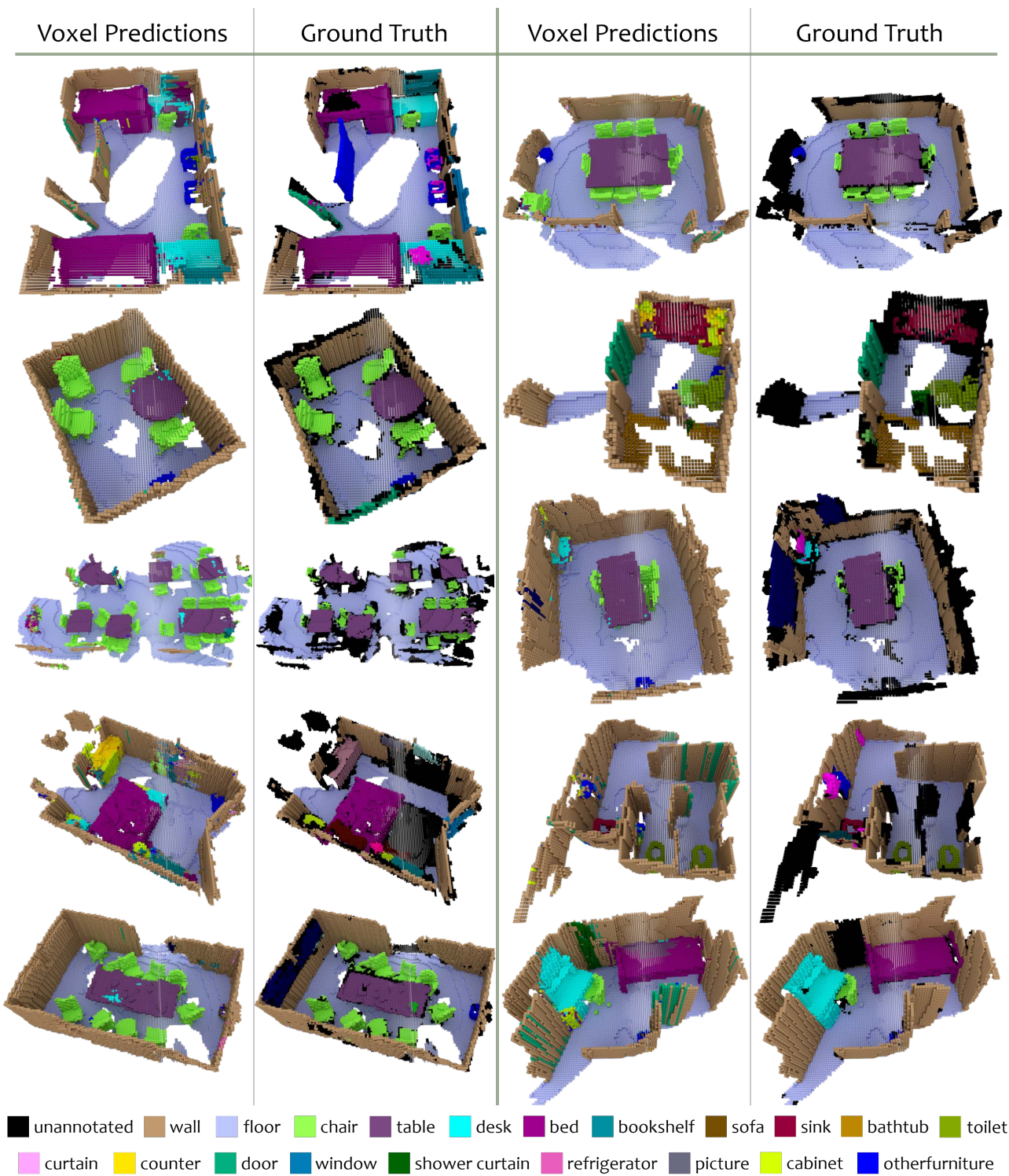

Figure 13. Semantic voxel labeling of 3D scans in ScanNet using our 3D CNN architecture. Voxel colors indicate predicted or ground truth category.

## C.2. Sensor Calibration

Sensor calibration is a critical, yet often overlooked part of RGB-D data acquisition. Our experiments showed that depth-to-color calibration is an important step in acquiring good 3D reconstructions from RGB-D sequences (see Fig. 15).

**Depth To Color Calibration** To align a depth image $\mathcal{D}$ to color image $\mathcal{C}$, we need to estimate intrinsic parameters of both sensors, the infrared camera $\mathbf{K}_\mathcal{D}$ and color camera $\mathbf{K}_\mathcal{C}$, as well as extrinsic transformation $\mathbf{T}_{\mathcal{D}\rightarrow\mathcal{C}}$. In our experiments we have found that using the set of intrinsic parameters of focal length, center of projection, and two barrel distortion coefficients models worked well for the used cameras. To obtain calibration parameters $\mathbf{K}_\mathcal{D}$ and $\mathbf{K}_\mathcal{C}$ we capture a series of color-infrared pairs showing an asymmetric checkerboard grid. We then estimate calibration parameters for each camera with Matlab's *CameraCalibrator* application. During this procedure we additionally obtain the world positions of calibration grid corners, and use them to estimate the transformation $\mathbf{T}_{\mathcal{D}\rightarrow\mathcal{C}}$.

**Depth Distortion Calibration** Previous work suggests that for consumer-level depth cameras there exists depth-dependent distortion that increases as camera moves away from the surface. Thus, we decided to augment our set of intrinsic parameters for depth cameras with a undistortion lookup table, as first suggested in Teichman et al. [84]. This look up table is a function $f(x,y,d)$, of spatial coordinates $x,y$ and observed depth $d$, returning a multiplication factor $m$ used to obtain undistorted depth $d' = md$. The table is computed from training pairs of observed and ground truth depths $d$ and $d_t$. However, unlike Teichman's unsupervised approach, which produces training pairs using carefully taken 'calibration sequences', we decided to design a supervised approach similar to that of Di Cicco [14]. However, we found that at large distances the depth distortion becomes so severe that approaches based on fitting planes to depth data are bound to fail. Thus to obtain training pairs $\{d, d_t\}$, we capture a color-depth video sequence of a large flat wall with a calibration target at the center, as the user moves away and towards the wall. To ensure successfull calibration process user needs to ensure that the viewed wall is the only observed surface and that it covers the entire field of view. With the captured color-depth sequence and previously estimated $\mathbf{K}_\mathcal{D}$, $\mathbf{K}_\mathcal{C}$, $\mathbf{T}_{\mathcal{D}\rightarrow\mathcal{C}}$ we can recover the the world positions of the calibration grid corners, effectively obtaining the ground truth plane locations for each of the captured depth images. For each pixel $x, y$ with depth $d$, we then shoot a ray through $x, y$ to intersect with the related plane. $d_t$ can be recovered from the point of intersection. The rest of our undistortion pipeline follows closely the that of Teichman et al. [84]. We found that undistorting depth images obtained by a Structure sensor leads to significantly improved tracking.

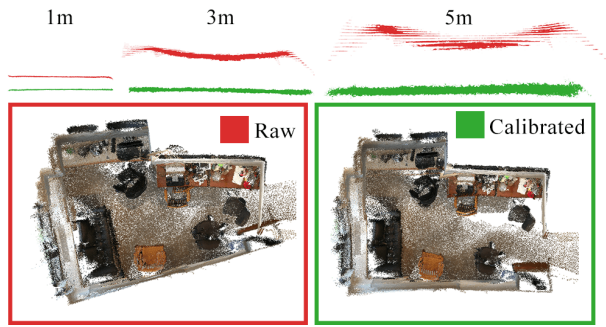

Figure 15. Comparison of calibration results. In the top row, we show results of calibration on a flat wall. As the distance increases the distortion becomes quite severe, motivating the need for depth distortion calibration. In the bottom row, we show results of frame-to-frame tracking on raw and calibrated data.

## C.3. Surface Reconstruction

Given a calibrated RGB-D sequence as input, a fused 3D surface reconstruction is obtained using the BundleFusion framework [12], as described in the main paper. The reconstruction is then cleaned by merging vertices within 1 mm of each other, and removing connected components with fewer than 7500 triangles. Following this cleanup step, two quadric edge collapse decimation steps are performed to produce lower triangle count versions of each surface mesh. Each decimation halves the number of triangles in the surface mesh, reducing the size of the original meshes from an average of 146 MB to 5.82 MB for the low resolution mesh. The mesh decimation step is important for reducing data transfer requirements and improving loading times during the crowdsourced annotation using our web-based UI.

## C.4. Crowdsourced Annotation UI

We deployed our semantic annotation task to crowd workers on the Amazon Mechanical Turk platform. Each annotation task began with an introduction (see Fig. 16) providing a basic overview of the task. The worker was then shown a reconstruction and asked to paint all object instances with a color and corresponding label. The worker was required to annotate at least 25% of the surface area of the reconstruction, and encouraged to cover at least 50%. Once the worker was done, they could submit by pressing a button. Workers were compensated with $0.50 for each annotation task performed.

The CAD model retrieval and alignment task began with a view of an already semantically annotated reconstruction and asked workers to click on objects to retrieve and place appropriate CAD models. Fig. 17 shows the initial instructions for an example reconstruction with several chairs. Workers for this task were required to place at least three objects before submitting. Once the worker was done, they were compensated with $1.00 for each completed task.

### C.5. Label cleaning and propagation

Labeling is performed on the surface mesh reconstruction, with several workers labeling each scan. To ensure that labels are consistent across workers, we use standard NLP techniques to clean up the labels. First, we use a manually curated list of good labels and their synonyms to compute a map to a single canonical label for each set, also including common misspellings by a small edit distance threshold of the given label. Labels with less than 5 counts are deemed unreliable and ignored in all statistics. Labels with more than 20 counts are manually examined and added to the list of good labels or collapsed as a synonym of a good label. The list of these frequent collapsed labels is also mapped to WordNet [18] synsets when possible, and to other common label sets that are commonly used for RGB-D and 3D CAD data (NYUv2 [58], ModelNet [91], and ShapeNetCore [6]).

Using the cleaned labels, we then compute an aggregated consensus labeling of each scene, since any individual crowdsourced annotation of a scene may not cover the entire scene, or may contain some errors. For each segment in the over-segmentation of a scene mesh, we first take the majority vote label. This groups together instances of the same class of objects, so we also compute a labeling purely based on geometric overlap; that is, we greedily take the unions of annotations which have $\geq 50\%$ overlap of segments. We then take the maximal intersections between these two labelings to obtain the final consensus.

After we have obtained the aggregated consensus semantic annotation for a scene, we then propagate these labels to the high-resolution mesh as well as to the 2D frames of the input RGB-D sequence. To propagate the labels to the high resolution mesh, we compute a kd-tree over the mesh vertices of the labeled coarse mesh, and we label each vertex of the high resolution mesh according to a nearest neighbor lookup in the kd-tree. We project the 3D semantic annotations to the input 2D frames by rendering the labeled mesh from the camera poses of each frame, and follow this with a joint dilation filter with the original RGB image and joint erosion filter with the original RGB image.

### C.6. Management UI

To enable scalability of our RGB-D acquisition and annotation, and continual transparency into the progress of scans throughout our framework, we created a web-based management UI to track and organize all data (see Fig. 19). When a user is finished scanning and presses the upload button on an iPad device, their scan data is automatically uploaded to our processing server, placed into a reconstruction queue, and immediately made visible in the management UI. As the reconstruction proceeds through the var-ious stages of data conversion, calibration, pose optimization and RGB-D fusion, alignment, cleanup, decimation, and segmentation, progress is visualized in the management UI. Thumbnail renderings of the generated surface reconstruction, and statistics such as total number of frames, reconstructed floor area etc. are automatically computed and can be used for filtering and sorting of the reconstructions. Similarly, during crowdsourced annotation, worker progress and aggregated annotated surface area statistics are visible and usable for sorting and filtering of the scan database.

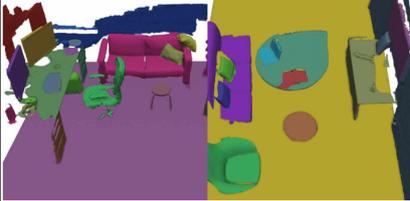

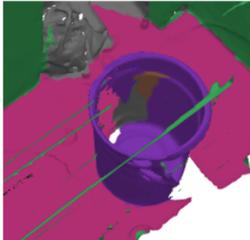

Figure 16. Instructions provided to crowd workers for our semantic annotation task. **Top:** instructions before the beginning of the task. **Bottom:** interface instructions during annotation.

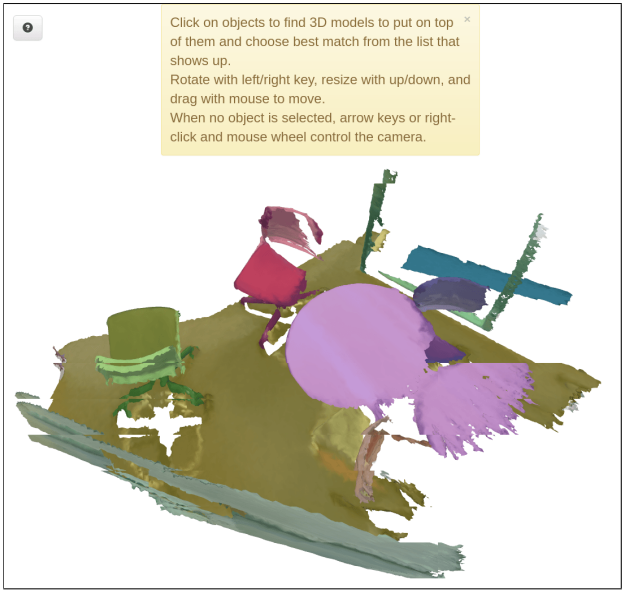

Figure 17. Instructions provided to crowd workers for our CAD model alignment task. The worker clicks on colored objects to retrieve and place CAD models.

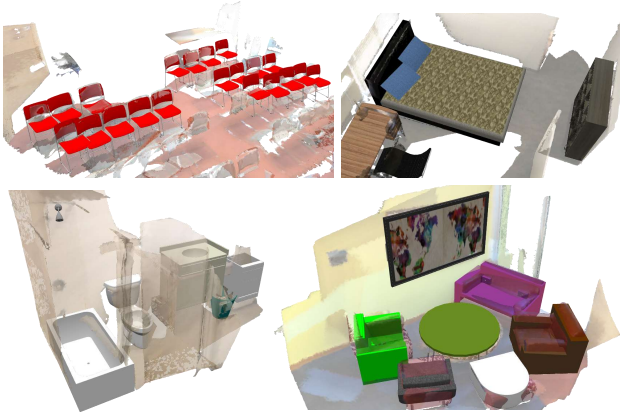

Figure 18. ShapeNetCore [6] CAD models retrieved and placed on ScanNet scans by crowd workers (scan mesh is transparent and CAD models are opaque). From top left clockwise: a classroom, bedroom, bathroom, and lounge scan.

Figure 19. Our web-based data management UI for ScanNet scan data.